\documentclass[format=sigconf]{acmart}

\usepackage{bm}
\usepackage{amsfonts}
\usepackage{amsmath}
\usepackage{color}
\usepackage{graphicx}
\usepackage{subfigure}
\usepackage{booktabs}
\usepackage{multirow}
\usepackage{enumitem}
\usepackage{setspace}
\usepackage{colortbl}
\usepackage{pifont}
\usepackage{balance}
\usepackage{array}
\usepackage{stfloats}

\newcommand{\bme}{\bm e}
\newcommand{\bmmu}{{\bm \mu}}
\newcommand{\bmx}{{\bm x}}

\newcommand{\model}{MotifClass}

\copyrightyear{2022}
\acmYear{2022}
\setcopyright{acmcopyright}\acmConference[WSDM '22]{Proceedings of the Fifteenth ACM International Conference on Web Search and Data Mining}{February 21--25, 2022}{Tempe, AZ, USA}
\acmBooktitle{Proceedings of the Fifteenth ACM International Conference on Web Search and Data Mining (WSDM '22), February 21--25, 2022, Tempe, AZ, USA}
\acmPrice{15.00}
\acmDOI{10.1145/3488560.3498384}
\acmISBN{978-1-4503-9132-0/22/02}

\begin{document}
\title{\textsc{\model}: Weakly Supervised Text Classification \\ with Higher-order Metadata Information} 
\author{Yu Zhang$^{1*}$, Shweta Garg$^{1*}$, Yu Meng$^{1}$, Xiusi Chen$^{2}$, Jiawei Han$^{1}$}
\affiliation{
\institution{$^1$Department of Computer Science, University of Illinois at Urbana-Champaign, IL, USA} 
\institution{$^2$Department of Computer Science, University of California, Los Angeles, CA, USA}
\institution{$^{1}$\{yuz9, shwetag2, yumeng5, hanj\}@illinois.edu, \ \ \ $^2$xchen@cs.ucla.edu}
}
\thanks{$^*$Equal Contribution.}

\begin{abstract}
We study the problem of weakly supervised text classification, which aims to classify text documents into a set of pre-defined categories with category surface names only and without any annotated training document provided. 
Most existing classifiers leverage textual information in each document. However, in many domains, documents are accompanied by various types of metadata (e.g., authors, venue, and year of a research paper). These metadata and their combinations may serve as strong category indicators in addition to textual contents. 
In this paper, we explore the potential of using metadata to help weakly supervised text classification. 
To be specific, we model the relationships between documents and metadata via a heterogeneous information network.
To effectively capture higher-order structures in the network,
we use motifs to describe metadata combinations. We propose a novel framework, named \textsc{\model}, which (1) selects category-indicative motif instances, (2) retrieves and generates pseudo-labeled training samples based on category names and indicative motif instances, and (3) trains a text classifier using the pseudo training data. 
Extensive experiments on real-world datasets demonstrate the superior performance of \textsc{\model} to existing weakly supervised text classification approaches.
Further analysis shows the benefit of considering higher-order metadata information in our framework.
\end{abstract}




\maketitle

\begin{spacing}{0.98}
\section{Introduction}
\begin{figure}
\centering
\subfigure[Two documents connecting with their metadata and text information]{
\includegraphics[width=0.45\textwidth]{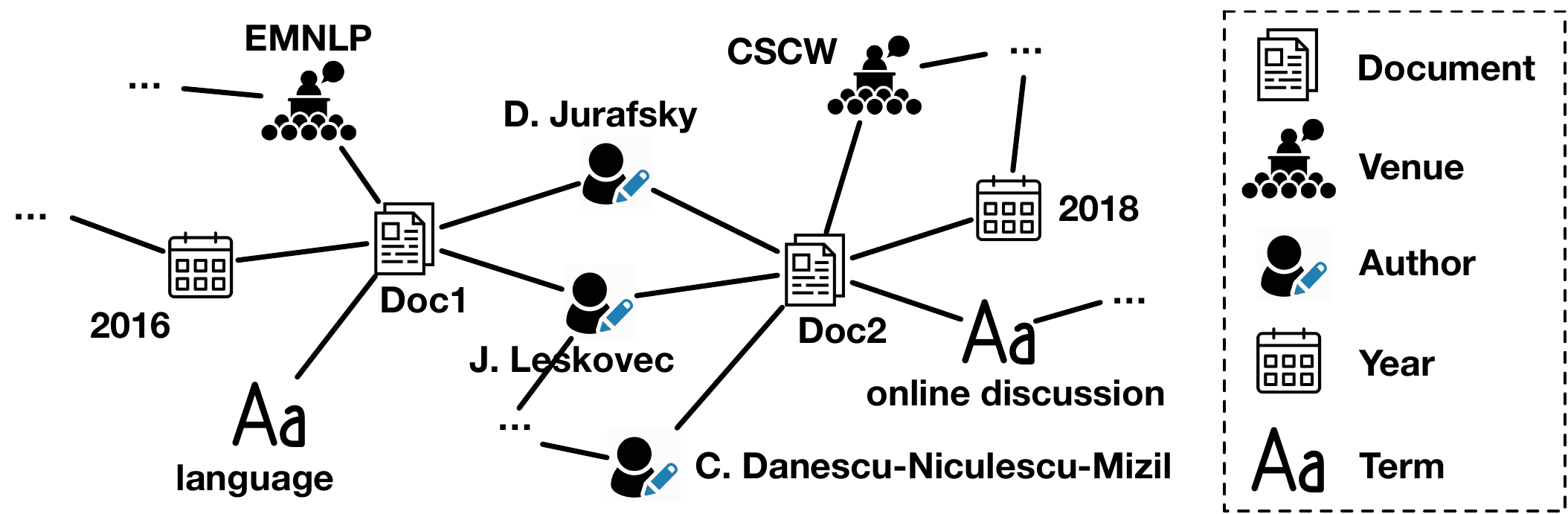}} \\
\vspace{-0.5em}
\subfigure[Examples of category-indicative and non-indicative motif instances associated with the two documents]{
\includegraphics[width=0.47\textwidth]{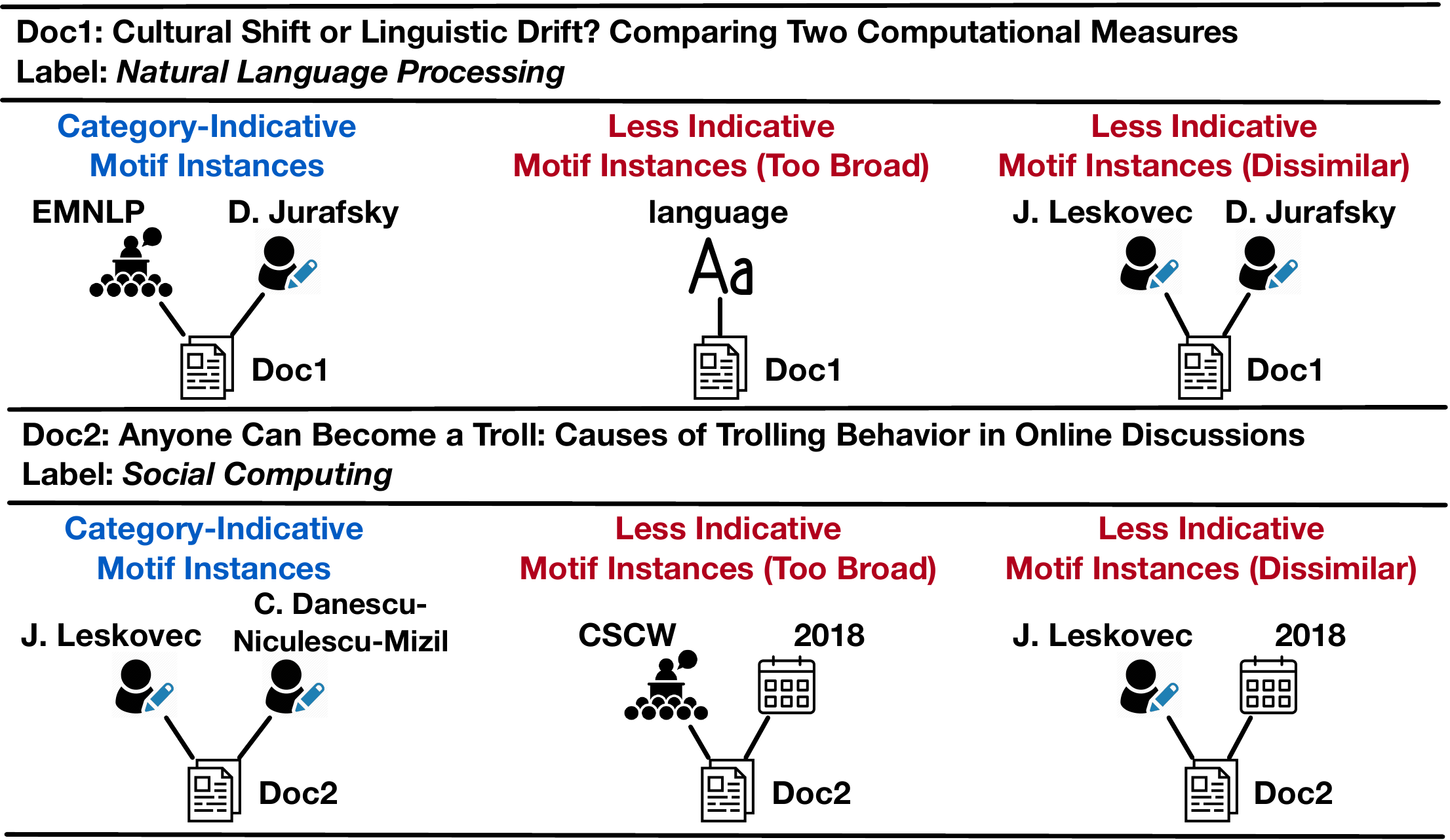}}
\vspace{-0.5em}
\caption{A network view of documents with metadata. Some metadata or metadata combinations (i.e., motifs) are category-indicative, while others are not.} 
\vspace{-0.5em}
\label{fig:intro}
\end{figure}

Text classification is a fundamental task in text mining with a wide spectrum of applications such as text geolocalization \cite{cheng2010you}, sentiment analysis \cite{pang2002thumbs}, and email intent detection \cite{wang2019context}. Following the routine of supervised learning, one can build a text classifier from human-annotated training documents. Many deep learning-based models (e.g., \cite{kim2014convolutional,yang2016hierarchical,guo2019star}) have achieved great performance in text classification when trained on a large-scale annotated corpus. Despite such a success, a frequent bottleneck of applying these models to a new domain is the acquisition of abundant annotated documents.
% people still find it challenging to apply these models to some real scenarios as the human labeling process is often very expensive and time-consuming.

Weakly supervised text classification, which relies on only category names or a few descriptive keywords to train a classifier, has recently gained increasing attention as it % automatically creates training data based on the weak supervision and 
eliminates the need for human annotations. 
Under the weakly supervised setting, most existing approaches leverage only the text data in each document \cite{chang2008importance,meng2018weakly,mekala2020contextualized,meng2020weakly,wang2021x}. However, in various domains, documents are beyond plain text sequences and are accompanied by different types of metadata (e.g., authors, venue, and year of a scientific paper; user and product of an e-commerce review). These metadata, together with text, provide better clues of the inter-relationship between multiple documents and thus are useful for inferring their categories.
% can connect documents together. 
Figure \ref{fig:intro}(a) provides a network view of an academic paper corpus with metadata. We can see that some metadata neighbors are helpful for predicting the category of a document. For example, the venue node \textit{EMNLP} suggests \textit{Doc1}'s relevance to Natural Language Processing.

Recent studies \cite{zhang2020minimally,mekala2020meta,zhang2021hierarchical} have confirmed that metadata signals are beneficial to weakly supervised text classification. However, in their work, the authors are less concerned with two important factors: higher-order metadata information and metadata specificity. 

\vspace{1mm}

\noindent \textbf{Higher-order Metadata Information\footnote{The term ``higher-order'' in this paper refers to higher-order network structures \cite{benson2016higher} represented by certain subgraph patterns \cite{shang2020nettaxo}, such as one document linked with two authors. It does not refer to multi-hop relationships or higher-order logic here.}.} Different types of metadata should be considered collectively in classification. For example, the combination of \textit{AAAI} and \textit{1990} is a strong indicator that the paper belongs to the traditional AI domain because the scope of AAAI was more focused in the early years. In comparison, either the venue or the year alone becomes a weaker signal. As another example, in Figure \ref{fig:intro}, \textit{Doc2} has two authors \textit{J. Leskovec} and \textit{C. Danescu-Niculescu-Mizil}. Neither of them alone is enough to predict the category Social Computing, but their co-authorship becomes category-indicative. Such higher-order information (called \underline{\textit{motifs}} in network science \cite{milo2002network,benson2016higher,shang2020nettaxo}) is not explored in \cite{zhang2020minimally,zhang2021hierarchical}. 

\vspace{1mm}

\noindent \textbf{Metadata Specificity.} To suggest the category of a document, a motif instance should be not only semantically close to that category but also specific enough to indicate only one category. For example, in Figure \ref{fig:intro}, the venue \textit{CSCW} may be linked with many papers related to Social Computing, but purely relying on \textit{CSCW} (or even the combination of \textit{CSCW} and \textit{2018}) will introduce noises because it is broader than the category. Similarly, the term \textit{language} in \textit{Doc1} is too broad to predict Natural Language Processing. Such metadata and text specificity is not considered in \cite{zhang2020minimally,mekala2020meta,zhang2021hierarchical}.

\vspace{1mm}

\noindent \textbf{Contributions.} In this paper, we study the problem of weakly supervised metadata-aware text classification. Being aware of higher-order metadata information and metadata specificity, we propose to discover discriminative and specific motifs for each category to help text classification. 
%Note that such motif discovery should be at the instance level (e.g., \textit{J. Leskovec} \& \textit{C. Danescu-Niculescu-Mizil}) instead of the type level (e.g., \textsc{Author} \& \textsc{Author}) because instances of the same motif type may have distinct abilities to indicate a category. 
Specifically, we propose \textsc{\model}, a framework that is built in three steps. (1) \textit{Indicative motif instance selection}: We leverage motif patterns (e.g., \textsc{Venue} \& \textsc{Author}) to obtain candidate motif instances (e.g., \textit{KDD} \& \textit{J. Leskovec}) in the dataset. Then, for each category, we select category-indicative motif instances based on their similarity with the label surface name as well as their specificity. To facilitate this, we propose a joint representation learning method to learn motif instance embedding and specificity simultaneously. (2) \textit{Pseudo-labeled training data collection}: By matching unlabeled documents with selected motif instances, we can retrieve documents that likely belong to a certain category. Besides retrieval, we propose to generate artificial training documents based on motif-aware text embeddings. The retrieval and the generation strategies are proved to be complementary to each other in creating pseudo training data. (3) \textit{Text classifier training}: We finally train a text classifier using collected pseudo training data. Note that our framework is compatible with any text classifier.

To summarize, this work makes the following contributions: 
\begin{itemize}[leftmargin=*]
    \item We propose a weakly supervised text classification model \textsc{\model}. It does not need any human-annotated document for training. Instead, it relies on category names and utilizes higher-order document metadata as additional supervision.
    \item We design an instance-level motif selection method to discover category-indicative metadata signals. The method is featured by a joint representation learning process that simultaneously learns the embedding and specificity of each motif instance.
    \item We conduct experiments on two real-world datasets to show the superiority of \textsc{\model} to existing weakly supervised metadata-aware text classification methods. 
\end{itemize}  

\section{Preliminaries}
\subsection{Text, Metadata, and Motif}
\begin{figure}
\centering
\includegraphics[width=0.47\textwidth]{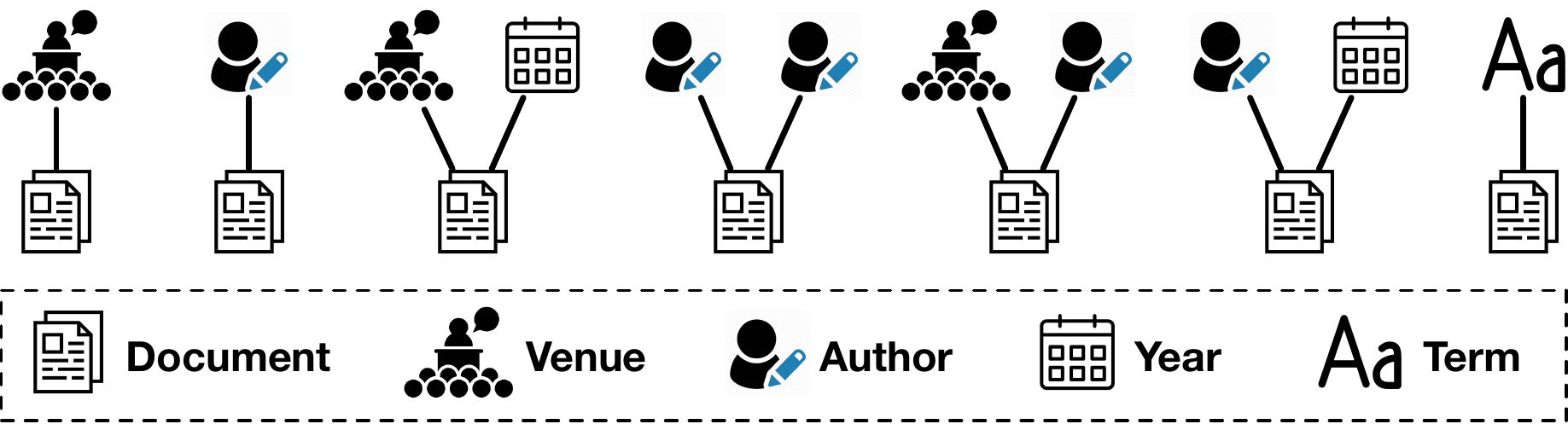}
\caption{Motif patterns used in an academic paper corpus.} 
\vspace{-0.5em}
\label{fig:motif1}
\end{figure}

\begin{figure*}
\centering
\includegraphics[width=0.97\textwidth]{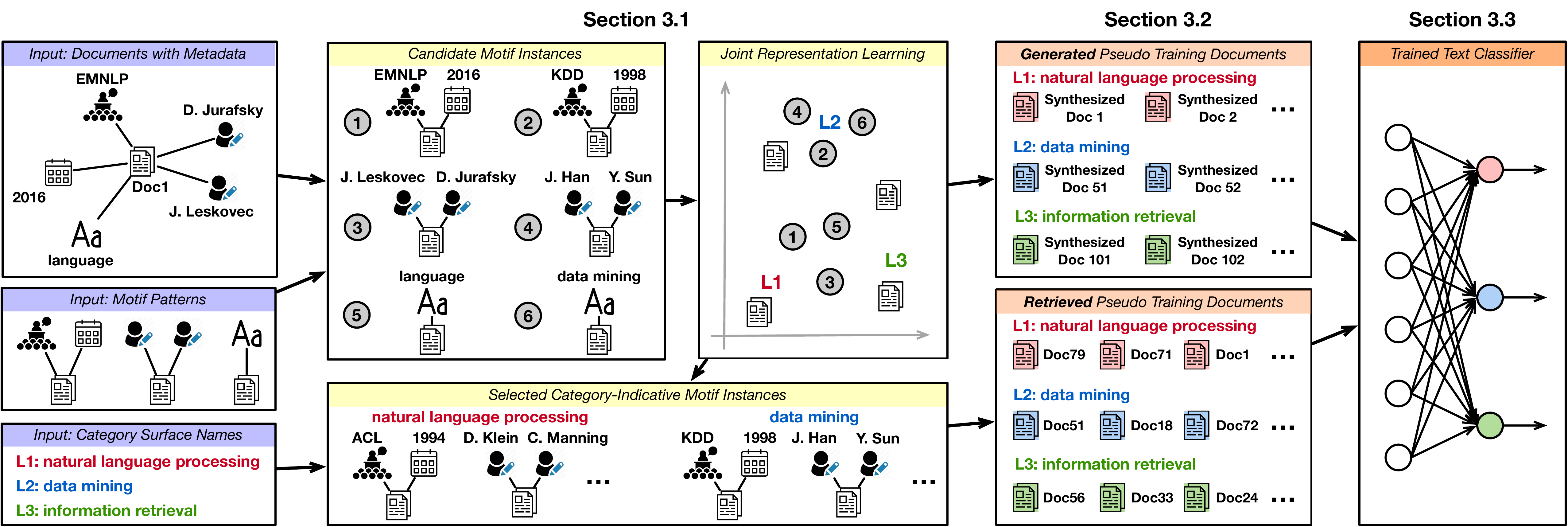}
\caption{The overview of our \textsc{\model} framework. We first discover category-indicative motif instances from documents with metadata. Pseudo-labeled training documents are then collected according to the selected motif signals. A text classifier is finally trained on the pseudo training data.} 
\vspace{-0.5em}
\label{fig:framework}
\end{figure*}

\noindent \textbf{Text.} We assume the text information of each document is a sequence of terms, denoted as $w_1w_2...w_N$. Each term $w_i$ here can be either a word or a phrase\footnote{We include phrases into our discussion because, in many scenarios, category names are not single words (e.g., Data Mining, Video Games). In practice, given a sequence of words, one can use existing phrase chunking tools \cite{manning2014stanford,shang2018automated} to detect phrases in it.}. To simplify our discussion, if a document has multiple text fields (e.g., title and abstract of a paper), we concatenate them into one sequence. 

\vspace{1mm}

\noindent \textbf{Metadata and Metadata Instance.} Documents are often associated with metadata \cite{zhang2020minimally,mekala2020meta}. For example, research papers can have \textsc{Author}, \textsc{Venue}, and \textsc{Year} fields. Each metadata type has its instances appearing in the dataset (e.g., \textsc{Venue}[\textit{EMNLP}], \textsc{Author}[\textit{D. Jurafsky}], \textsc{Year}[\textit{2016}]). 

As illustrated in Figure \ref{fig:intro}, we can construct a heterogeneous information network (HIN) \cite{sun2012mining} to describe documents with metadata. The formal definition of HIN is as follows.

\begin{definition}{(Heterogeneous Information Network \cite{sun2012mining})} 
An HIN is a graph $G = (\mathcal{V}, \mathcal{E})$ with a node type mapping $\phi: \mathcal{V}\rightarrow\mathcal{T}_\mathcal{V}$ and an edge type mapping $\psi: \mathcal{E}\rightarrow\mathcal{T}_\mathcal{E}$. Either the number of node types $|\mathcal{T}_\mathcal{V}|$ or the number of edge types $|\mathcal{T}_\mathcal{E}|$ is larger than 1.
\label{def:hin}
\end{definition}

In our constructed HIN, $\mathcal{V}$ consists of document nodes, term nodes, and all metadata instances; $\mathcal{E}$ includes edges connecting each document with its metadata information and words/phrases.

\vspace{1mm}

\noindent \textbf{Motif Pattern and Motif Instance.} In an HIN, a motif pattern refers to a subgraph at the type level.

\begin{definition}{(Motif Pattern \cite{shang2020nettaxo})} 
A motif pattern in an HIN is a connected graph $p$. Each node in $p$ is a node type $\in \mathcal{T}_\mathcal{V}$, and each edge in $p$ is an edge type $\in \mathcal{T}_\mathcal{E}$.
\label{def:motif}
\end{definition}

In the document classification task, we focus on motif patterns with one \textsc{Document} node. In this way, motif patterns essentially describe the semantics of metadata and their combinations. For example, Figure \ref{fig:motif1} shows the motif patterns that can be used in an academic paper corpus. They are able to model the relationship between documents and (higher-order) metadata. For ease of notation, in this paper, when representing a motif pattern, we omit the \textsc{Document} node and write the metadata node(s) only. For example, in Figure \ref{fig:motif1}, the third pattern from the left can be written as \textsc{Venue}-\textsc{Year}, and the fourth one can be written as \textsc{Author}-\textsc{Author}. We view the connection between a document and a term also as a motif pattern (i.e., \textsc{Term}), so that we can describe text information.

Similar to a single metadata type, a motif pattern has its instances (e.g., \textsc{Venue}[\textit{EMNLP}]-\textsc{Year}[\textit{2016}], \textsc{Term}[\textit{data mining}]). Note that each word/phrase $w_i$ appearing in the corpus will be viewed as a motif instance \textsc{Term}[$w_i$].

\subsection{Problem Definition}
Given a collection of documents $\mathcal{D}=\{d_i\}_{i=1}^{|\mathcal{D}|}$ and a set of categories $\mathcal{L}=\{l_j\}_{j=1}^{|\mathcal{L}|}$, the text classification task is to assign a category label $l_j$ to each document $d_i$. In this paper, we study the \textit{weakly supervised} setting, where no human-annotated training data is needed, and the only descriptive signal of each category is the surface text of its category name. Each category $l$ has only one surface name (a term, denoted as $n_l$).
We allow the category name to be either a single word (e.g., database) or a phrase (e.g., data mining). This assumption is more relaxed than that of previous studies using BERT-based models \cite{mekala2020contextualized,meng2020weakly,wang2021x}, where the category name must be a single word in the vocabulary of BERT.

Following the setting in \cite{mekala2020meta,shang2020nettaxo}, we ask users to specify a set of possibly useful motif patterns $\mathcal{P}=\{p_i\}_{i=1}^{|\mathcal{P}|}$ as input to our model, like what we show in Figure \ref{fig:motif1}. (This setting is aligned with many HIN embedding studies \cite{dong2017metapath2vec,wang2019heterogeneous,yang2020heterogeneous} that prefer users to input a set of meta-paths.) If users do not have such prior knowledge, we can also enumerate all possible metadata combinations (including those with $\geq 3$ metadata nodes) as candidate motif patterns.
Note that our framework automatically refines motif signals through instance-level selection, thus is robust to the existence of unreliable input motif patterns. 

To summarize, our problem definition is as follows.
\begin{definition}{(Problem Definition)} 
Given a set of unlabeled documents $\mathcal{D}$ with metadata, a label space $\mathcal{L}$, the surface name of each category $\{n_l: l\in \mathcal{L}\}$, and a set of candidate motif patterns $\mathcal{P}$, the task is to assign a label $l \in \mathcal{L}$ to each $d \in \mathcal{D}$.
\label{def:problem}
\end{definition}

\section{Framework}
Figure \ref{fig:framework} illustrates the overall \textsc{\model} framework. The core idea is to use category names and higher-order metadata information to create pseudo-labeled training data. To implement this idea, we first discover category-indicative motif instances for each category through a joint representation learning process (\textbf{Section \ref{sec:method1}}). Then, we retrieve and generate pseudo-labeled training documents based on selected motif instances and learned motif-aware embeddings, respectively (\textbf{Section \ref{sec:method2}}). Finally, using pseudo-labeled documents, we train a text classifier (\textbf{Section \ref{sec:method3}}).

\subsection{Selecting Indicative Motif Instances}
\label{sec:method1}
Given the candidate motif patterns, we first find all instances of these motifs in the corpus $\mathcal{D}$ (instances with frequency below a certain threshold will be discarded). We denote the set of candidate motif instances as $\mathcal{M}=\{m_i\}_{i=1}^{|\mathcal{M}|}$. For each category $l \in \mathcal{L}$, our first step is to select a group of category-indicative motif instances $\mathcal{M}_l \subseteq \mathcal{M}$. We assume the category name \textsc{Term}[$n_l$] must be an indicative motif instance of $l$. Then the goal is to find other indicative instances according to \textsc{Term}[$n_l$]. We propose the following two criteria.

\vspace{1mm}

\noindent \textbf{Similarity.} The selected motif instance should be semantically similar to the corresponding category name. In other words, if we embed all motif instances into the same latent space, we expect each selected motif instance (e.g., \textsc{Venue}[\textit{EMNLP}]-\textsc{Author}[\textit{D. Jurafsky}]) to have high embedding cosine similarity with the category name (e.g., \textsc{Term}[\textit{natural language processing}]).

\vspace{1mm}

\noindent \textbf{Specificity.} The selected motif instances should not indicate multiple categories at the same time (e.g., \textsc{Venue}[\textit{AAAI}]), so that we can use these instances to infer high-quality pseudo training data for each category. To facilitate this, we require the selected motif instances to be semantically more specific than the category name.

\subsubsection{Joint Embedding and Specificity Learning}
Based on these two requirements, for each motif instance $m \in \mathcal{M}$, we propose to learn two parameters $\bme_m$ and $\kappa_m$ from the corpus. Here, $\bme_m$ is the embedding vector of $m$, and $\kappa_m$ is the specificity of the instance (a scalar). The larger $\kappa_m$ is, the more focused semantics the instance should indicate. For example, we should expect $\kappa_{\small \textsc{Venue}[\textit{AAAI}]} < \kappa_{\small \textsc{Venue}[\textit{EMNLP}]} < \kappa_{\small \textsc{Author}[\textit{D. Jurafsky}]}$.

To simultaneously estimate $\bme_m$ and $\kappa_m$, we propose a joint representation learning process that embeds motif instances, categories, and documents into the same latent space. It considers the following two types of proximity in the learning objective.

\vspace{1mm}

\noindent \textbf{Motif Instance--Document Proximity.} Previous studies on word embedding \cite{tang2015pte,meng2020cate} encourage the proximity between each word and its belonging document. This idea can be directly generalized from words to motif instances. Given an instance $m$ and a document $d$, where $m$ appears in $d$ \footnote{We say a motif instance $m$ appears in a document $d$ if and only if $d$ contains all metadata instances of $m$. For example, in Figure \ref{fig:intro}, the instance \textsc{Venue}[\textit{EMNLP}]-Author[\textit{D. Jurafsky}] appears in \textit{Doc1}.}, we aim to maximize the following probability:
\begin{equation}
    p(d|m) = \frac{\exp(\kappa_m \bme_m^T\bme_d)}{\sum_{d'\in \mathcal{D}}\exp(\kappa_m \bme_m^T\bme_{d'})}.
\label{eqn:motif-document}
\end{equation}
If we ignore $\kappa_m$, Eq. (\ref{eqn:motif-document}) is essentially a softmax function widely used in embedding learning. \citet{meng2020cate} first introduce ``$\kappa$'' into the softmax function to model word specificity. We extend their technique to the motif case. To explain why $\kappa_m$ can represent the specificity of $m$, we follow \cite{meng2020cate} and introduce the von Mises-Fisher (vMF) distribution \cite{fisher1953dispersion}.

\begin{definition}{(The Von Mises-Fisher Distribution \cite{fisher1953dispersion})} 
The vMF distribution is defined on a unit sphere $\mathbb{S}^{\delta-1}=\{\bmx \in \mathbb{R}^\delta: ||\bmx||_2=1\}$. It is parameterized by the mean direction vector $\bmmu$ and the concentration parameter $\kappa$. The probability density function is
\begin{equation}
    {\rm vMF}(\bmx|\bmmu, \kappa) = c_{\delta}(\kappa) \exp(\kappa \bmmu^T\bmx),
\end{equation}
where $\bmx \in \mathbb{S}^{\delta-1}$, $\bmmu \in \mathbb{S}^{\delta-1}$, and $\kappa \geq 0$. Here, $c_{\delta}(\kappa)$ is a constant related to $\kappa$ and $\delta$ only.
\label{def:vmf}
\end{definition}

\noindent Intuitively, the vMF distribution can be viewed as an analogue of the Gaussian distribution on a sphere. The distribution concentrates around $\bmmu$, and is more concentrated if $\kappa$ is larger. 

Motivated by the fact that directional similarity is more effective in capturing semantics \cite{meng2019spherical,meng2020cate}, we require all embedding vectors $\bme_m$ and $\bme_d$ in Eq. (\ref{eqn:motif-document}) to reside on a unit sphere, then we have
\begin{equation}
\begin{split}
\lim_{|\mathcal{D}|\rightarrow\infty} p(d|m) &= \frac{\exp(\kappa_m \bme_m^T\bme_d)}{\int_{\mathbb{S}^{\delta-1}}\exp(\kappa_m \bme_m^T\bme_{d'}){\rm d}\bme_{d'}} = \frac{\exp(\kappa_m \bme_m^T\bme_d)}{1/c_{\delta}(\kappa_m)} \\
&= c_{\delta}(\kappa_m) \exp(\kappa_m \bme_m^T\bme_d) = {\rm vMF}(\bme_d|\bme_m, \kappa_m).
\end{split}
\label{eqn:vmf}
\end{equation}
The second step holds because $\int_{\mathbb{S}^{\delta-1}} c_{\delta}(\kappa_m) \exp(\kappa_m \bme_m^T\bme_{d'}){\rm d}\bme_{d'} = \int_{\mathbb{S}^{\delta-1}} {\rm vMF}(\bme_{d'}|\bme_m, \kappa_m) {\rm d}\bme_{d'} \equiv 1$.
% (i.e., the probability density function always integrates to 1 over the whole space.) 
Eq. (\ref{eqn:vmf}) essentially assumes that given the motif instance $m$, the embeddings of documents containing $m$ are generated from ${\rm vMF}(\cdot|\bme_m, \kappa_m)$. For a motif instance with more general meaning (e.g., \textsc{Venue}[\textit{AAAI}]), it will appear in more diverse documents. Therefore, its learned vMF distribution will have a lower concentration parameter $\kappa_m$ than that of a more specific instance (e.g., \textsc{Venue}[\textit{EMNLP}]). This explains why $\kappa_m$ can represent the specificity of $m$.

Given the probability $p(d|m)$ in Eq. (\ref{eqn:motif-document}), we aim to maximize the log-likelihood
\begin{equation}
\mathcal{J}_{\rm Doc} = \sum_{m \in \mathcal{M}} \ \ \sum_{d:\ m \text{ appears in } d} \log p(d|m).
\end{equation}

\vspace{1mm}

\noindent \textbf{(\textsc{Term}) Motif Instance--Context Proximity.} Words/phrases have local context information. To be specific, given a text sequence $w_1w_2...w_N$, \citet{mikolov2013distributed} define the local context of $w_i$ as $\mathcal{C}(w_i, h)=\{w_j: i-h \leq j \leq i+h, j\neq i\}$, where $h$ is the context window size. 
We view each word/phrase as a \textsc{Term} instance, so the local context of a \textsc{Term} motif instance can be written as $\mathcal{C}({\textsc{Term}}[w_i], h)=\{{\textsc{Term}}[w_j]: i-h \leq j \leq i+h, j\neq i\}$. According to the Skip-Gram model \cite{mikolov2013distributed}, we consider the following proximity.
\begin{equation}
p(\mathcal{C}(m, h)|m) = \prod_{m_+ \in \mathcal{C}(m, h)} \frac{\exp(\kappa_{m}\bme_{m}^T\bme_{m_+})}{\sum_{m_-}\exp(\kappa_{m}\bme_{m}^T\bme_{m_-})}.
\end{equation}
Similar to Eq. (\ref{eqn:motif-document}), the specificity $\kappa_m$ is added into the softmax function. The log-likelihood is given by
\begin{equation}
    \mathcal{J}_{\rm Ctxt} = \sum_{d \in \mathcal{D}} \ \ \sum_{m:\ \text{ \textsc{Term} instance, appears in } d} \log p(\mathcal{C}(m, h)|m).
\label{eqn:local}
\end{equation}

Based on the two types of proximity, our joint representation learning process can be cast as an optimization problem:
\begin{equation}
\max_{\bme_m, \bme_d, \kappa_m} \mathcal{J} =  \mathcal{J}_{\rm Doc} + \mathcal{J}_{\rm Ctxt}, \ \ \  \text{s.t.} \ \ \bme_m, \bme_d \in \mathbb{S}^{\delta-1},\ \kappa_m \geq 0.
\label{eqn:obj}
\end{equation}
The optimization process of this objective can be found in Appendix \ref{sec:app_opt}.

\subsubsection{Motif Instance Selection}
After obtaining the embedding vector and specificity of each motif instance, we are able to select a set of indicative motif instances $\mathcal{M}_l$ for each category. First, we assume the category name $n_l$ must be indicative, so we have the \textsc{Term} instance $m_l={\textsc{Term}}[n_l]$ in $\mathcal{M}_l$. Then, we find top-ranked instances and add them into $\mathcal{M}_l$. The ranking criterion is
\begin{equation}
    \max_{m \in \mathcal{M}} \cos(\bme_m, \bme_{m_l}),\ \ \text{where } \kappa_m \geq \eta \cdot \kappa_{m_l}.
\end{equation}
Here, $\eta > 1$ is a hyperparameter. Intuitively, from all motif instances that are more specific than the category name (i.e., the \textit{specificity} criterion), we select a number of instances closest to the category name in the embedding space (i.e., the \textit{similarity} criterion).

\subsection{Retrieving and Generating Pseudo-Labeled Training Data}
\label{sec:method2}
Based on the selected motif instances and motif-aware embeddings, we aim to collect pseudo-labeled training data $\mathcal{D}_l$ for each category $l$. In this paper, we propose two ways, \textit{retrieval} and \textit{generation}. The idea of retrieval is to use category-indicative motif instances $\mathcal{M}_l$ to find existing unlabeled documents which likely belong to $l$. In contrast, the idea of generation is to generate artificial documents (i.e., sequences of text and metadata) that have close meaning to $l$.

\vspace{1mm}

\noindent \textbf{Retrieval.} Given a document $d \in \mathcal{D}$ and a category $l \in \mathcal{L}$, we calculate the score that $d$ belongs to $l$ by counting the number of $l$'s indicative motif instances appearing in $d$. Formally,
\begin{equation}
    {\rm score}(d,l) = \sum_{m \in \mathcal{M}_l} \textbf{1}(m \text{ appears in } d).
\end{equation}
Here, $\textbf{1}(\cdot)$ is the indicator function. 
% Instead of treating all selected instances equally, we have also tried to weight them by specificity (e.g., $\kappa_m$ or $\frac{1}{\kappa_m}$) when calculating the score. However, such modification does not bring performance improvement.

For each category $l$, we retrieve a set of documents $\mathcal{D}^{\rm R}_l \subseteq \mathcal{D}$ as the pseudo training data with label $l$. The retrieved documents should have a high score with $l$ and a score of 0 with any other category. In other words, the ranking criterion is
\begin{equation}
    \max_{d \in \mathcal{D}} {\rm score}(d,l),\ \ \text{where } {\rm score}(d,l')=0\ \ \  (\forall l'\neq l).
\label{eqn:retr}
\end{equation}
By Eq. (\ref{eqn:retr}), top-ranked documents are selected and added to $\mathcal{D}^{\rm R}_l$.

\vspace{1mm}

\noindent \textbf{Generation.} Given $l \in \mathcal{L}$, we generate a set of synthesized documents $\mathcal{D}^{\rm G}_l$ that belong to the category $l$. To generate text related to a certain topic, we follow the idea in \cite{zhang2020minimally} and leverage our joint representation learning space. Specifically, there are two major steps: \underline{\textit{Step 1}}: given a category, generate a document embedding $\bme_d$ that is semantically close to the category. \underline{\textit{Step 2}}: given the document embedding $\bme_d$, generate a sequence of metadata instances and words/phrases that are coherent with the document semantics.

\underline{\textit{Step 1}}: We have obtained the category name embedding $\bme_{m_l}$ in the joint representation learning step. When generating the document embedding, we expect $\bme_d$ to be close to  $\bme_{m_l}$ in the embedding space, thus we adopt the vMF distribution.
\begin{equation}
    \bme_d \sim {\rm vMF}(\cdot|\bme_{m_l}, \kappa).
\end{equation}
Note that we cannot use a softmax function here because we are ``creating'' a new document instead of sampling one from the existing pool. Therefore, we use the vMF distribution which, according to Eq. (\ref{eqn:vmf}), is a good approximation of a softmax function.

\underline{\textit{Step 2}}: Now, to form a complete document, we aim to generate a sequence of metadata instances and words/phrases. Note that most words/phrases and metadata instances appearing in the dataset can be represented as a motif instance (e.g., \textsc{Venue}[\textit{EMNLP}], \textsc{Author}[\textit{D. Jurafsky}], \textsc{Term}[\textit{language}]).
We have learned the embeddings of all motif instances above a certain frequency threshold. Based on these embeddings, the probability of generating a word/phrase or metadata instance $m$ in a document $d$ is given by
\begin{equation}
    p(m|\bme_d) = \frac{\exp(\bme_d^T\bme_{m})}{\sum_{m' \in \mathcal{M}_{\rm Gen}} \exp(\bme_d^T\bme_{m'})}\ \ \ (\forall m \in \mathcal{M}_{\rm Gen}).
\label{eqn:gen}
\end{equation}
Here, $\mathcal{M}_{\rm Gen}$ is the set of words/phrases and metadata instances used to generate $d$. In practice, we set $\mathcal{M}_{\rm Gen}$ to be the top-50 nearest neighbors of $\bme_d$ in the embedding space. We do not use all words/phrases and metadata instances in the embedding space because the computational cost of $\sum_{m' \in \mathcal{M}_{\rm Gen}} \exp(\bme_d^T\bme_{m'})$ will be very high in that case. Using Eq. (\ref{eqn:gen}) repeatedly, we can obtain a sequence of metadata instances and words/phrases $m_1m_2...m_N$.

The final set of pseudo-labeled training documents $\mathcal{D}_l$ is the union of the retrieved ones $\mathcal{D}^{\rm R}_l$ and the generated ones $\mathcal{D}^{\rm G}_l$. We use the combination of retrieval and generation strategies because they have different merits. Retrieved documents are real, thus have higher linguistic quality. However, the input corpus $\mathcal{D}$ may not have lots of documents whose pseudo-label predictions are confident enough. In contrast, generated documents are artificial, but the number of generated documents is not limited by the size of $\mathcal{D}$.

\subsection{Training a Text Classifier}
\label{sec:method3}
Our framework is compatible with any text classification model as a classifier (e.g., CNN \cite{kim2014convolutional}, HAN \cite{yang2016hierarchical}, Transformer \cite{vaswani2017attention}). The goal of this paper is not to develop a novel classifier. Therefore, following previous studies \cite{meng2018weakly,zhang2020minimally,zhang2021hierarchical}, we adopt Kim-CNN \cite{kim2014convolutional} as our classifier, with all parameter settings the same as those in \cite{meng2018weakly}.
% We use filters with widths 2, 3, 4, and 5. For each width, we generate 20 feature maps. After pooling, the features are passed through a fully connected softmax layer whose output is the probability distribution over labels.

Given a pseudo-labeled training document $d \in \mathcal{D}_l$, we feed both its text and metadata information into the classifier. Specifically, if $d$ is retrieved, we concatenate its metadata and text information into one sequence. For example, given the paper \textit{Doc1} in Figure \ref{fig:intro}, the input sequence is

\vspace{1mm}

\noindent ``\textsc{Author}[\textit{W. Hamilton}] \textsc{Author}[\textit{J. Leskovec}] \textsc{Author}[\textit{D. Jurafsky}] \textsc{Venue}[\textit{EMNLP}] \textsc{Year}[\textit{2016}] \textit{cultural shift or linguistic drift} ...''

\vspace{1mm}

\noindent If $d$ is generated, it already has a mixed sequence of metadata instances and words/phrases. 
We train Kim-CNN on each $d$ with its pseudo-label. The training loss is the negative log-likelihood. The initialized word/phrase and metadata embeddings are those learned in joint representation learning (i.e., $\bme_m$) since most words/phrases or metadata instances can be viewed as a motif instance.

We would like to report that we have also tried BERT \cite{devlin2019bert} as our classifier, but its performance is not so good as that of Kim-CNN, possibly because the fixed vocabulary of BERT restricts its capacity to deal with metadata instances (e.g., author names, product IDs).

\section{Experiments}
\subsection{Setup}
\noindent \textbf{Datasets.} We use two real-world datasets from different domains for evaluation\footnote{The code and datasets are available at \\ \url{https://github.com/yuzhimanhua/MotifClass}.}. Dataset statistics are listed in Table \ref{tab:data}.

\begin{table}[]
\caption{Dataset statistics.}
\vspace{-0.5em}
\scalebox{0.82}{
\begin{tabular}{c|>{\centering\arraybackslash}p{3.5cm}>{\centering\arraybackslash}p{3cm}}
\hline
                  & \textbf{MAG-CS \cite{zhang2021match}}                                                                                                            & \textbf{Amazon \cite{mcauley2013hidden}}                                                                      \\ \hline
\#Documents       & 203,157                                                                                                           & 100,000                                                                     \\ 
Avg Doc Length    & 125                                                                                                               &     120                                                                        \\ 
\#Categories      & 20                                                                                                                & 10                                                                          \\ 
Text Fields         & title, abstract                                                                                               & headline, review                                                              \\ 
Metadata Fields         & \textsc{Author}, \textsc{Venue}, \textsc{Year}                                                                                               & \textsc{User}, \textsc{Product}                                                               \\ \hline
\end{tabular}
}
\label{tab:data}
\end{table}

\begin{figure}
\centering
\includegraphics[width=0.47\textwidth]{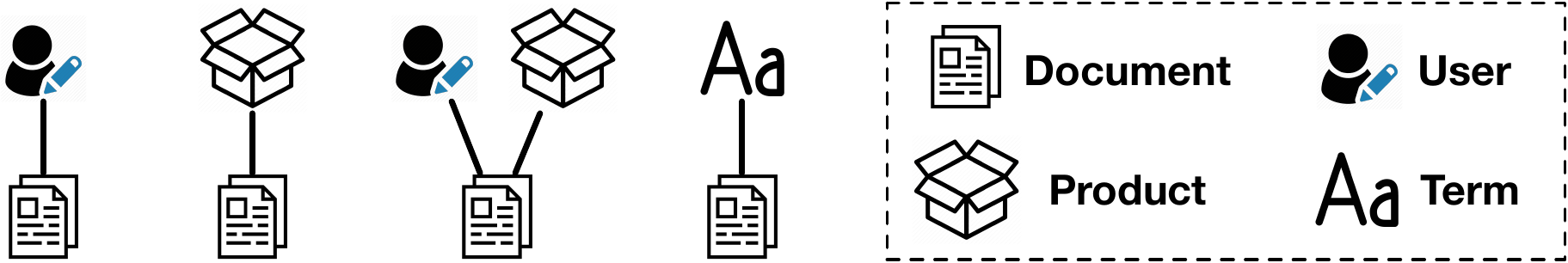}
\vspace{-0.5em}
\caption{Motif patterns used in the Amazon dataset.} 
\vspace{-0.5em}
\label{fig:motif2}
\end{figure}

\begin{itemize}[leftmargin=*]
    \item \textbf{MAG-CS \cite{zhang2021match}} is constructed from the Microsoft Academic Graph (MAG). It consists of papers published in 105 top CS conferences from 1990 to 2020. Each paper has labels at different levels of the MAG taxonomy. We use labels in the highest level for classification, and we remove papers that belong to two or more categories because our problem setting is single-label classification. The candidate motif patterns are listed in Figure \ref{fig:motif1}.
    \item \textbf{Amazon \cite{mcauley2013hidden}} is a crawl of Amazon product reviews. Each review is associated with its user (i.e., reviewer) and product IDs. 10 large categories are selected and 10,000 reviews are sampled from each category. The used motif patterns are listed in Figure \ref{fig:motif2}.
\end{itemize}

\noindent In MAG-CS, phrase terms are already recognized. In Amazon, we use AutoPhrase \cite{shang2018automated} to detect phrases in the text. The whole datasets are used for evaluation because our weakly supervised setting does not require any training document with ground-truth labels.

\vspace{1mm}

\noindent \textbf{Compared Methods.} We compare \textsc{\model} with the following methods, including both weakly supervised text classification approaches and HIN embedding methods.

\begin{itemize}[leftmargin=*]
    \item \textbf{WeSTClass \cite{meng2018weakly}} is a weakly supervised text classification approach. It can take category surface names as supervision and applies a pre-training and self-training scheme.
    \item \textbf{WeSTClass+Metadata} is an easy extension of WeSTClass. Since WeSTClass considers text information only, we concatenate all metadata instances of a document with its text as the input sequence, so that WeSTClass can utilize metadata signals.
    \item \textbf{MetaCat \cite{zhang2020minimally}} is a weakly supervised metadata-aware text classification approach. It takes a small set of labeled documents, instead of category names, as supervision. To align the experiment setting, we use the pseudo-labeled documents retrieved by \textsc{\model} (i.e., $\mathcal{D}_l^{\rm R}$) as the supervision of MetaCat.
    % \footnote{Because the supervision format in this paper is different from that in the original MetaCat paper \cite{zhang2020minimally}, the scores of MetaCat in Table \ref{tab:exp_results} are also different from those reported in \cite{zhang2020minimally} on the same Amazon dataset.}
    \item \textbf{META \cite{mekala2020meta}} is a weakly supervised metadata-aware text classification approach. It can take category names as supervision and iteratively performs classification and motif instance expansion.
    \item \textbf{LOTClass \cite{meng2020weakly}} is a weakly supervised text classification approach based on BERT. It takes category names or keywords as supervision, but each category name/keyword must be a single word in the vocabulary of BERT. 
    To apply it here, we separate each phrase category name into single words and remove those common words that appear in multiple category names.
    \item \textbf{Metapath2Vec \cite{dong2017metapath2vec}} is an HIN embedding method. We use it to embed terms, documents, and metadata instances into the same space.
    The category of each document is given by its nearest category name in the embedding space.
    \item \textbf{HIN2Vec \cite{fu2017hin2vec}} is an HIN embedding method that considers meta-path embeddings in addition to node embeddings. 
    % It does not need pre-specified meta-paths. 
    We perform nearest neighbor search after learning the embeddings to classify each document.
    \item \textbf{HGT \cite{hu2020heterogeneous}} is a recent heterogeneous graph neural network model. We adopt the unsupervised unattributed version of HGT and perform nearest neighbor search after learning node embeddings.
    \item \textbf{\textsc{\model}-NoHigherOrder} is an ablation version of \textsc{\model} that does not leverage higher-order metadata information. Specifically, it only considers single metadata types (e.g., \textsc{Venue}, \textsc{Author}, and \textsc{Term} in Figure \ref{fig:motif1}) as input motifs.
    \item \textbf{\textsc{\model}-NoSpecificity} is another ablation version of \textsc{\model} that does not consider specificity of motif instances. In other words, for each $m$, $\kappa_m$ is fixed to be 1. There is no specificity requirement when selecting motif instances.
\end{itemize}

\noindent We also present the performance of BERT \cite{devlin2019bert} under the fully supervised setting (shown as \textbf{Supervised BERT} in Table \ref{tab:exp_results}), where we perform a random 80\%-10\%-10\% train-dev-test split of the datasets.

% There are other BERT-based weakly supervised text classification methods \cite{mekala2020contextualized,wang2021x}. Again, they require the category names to be single words, which is not the case in both of our datasets. Since we already include one recent BERT-based method -- LOTClass \cite{meng2020weakly} -- as our baseline, we do not include more with the same drawback.

\vspace{1mm}

\noindent \textbf{Implementation and Hyperparameters.} We discard infrequent motif instances that appear in less than 5 documents. The embedding dimension $\delta=100$. The context window size $h=5$. During embedding learning, for each positive sample, we collect 5 negative samples. The size of selected motif instances $|\mathcal{M}_l|=50$. The specificity criterion is $\kappa_m \geq 2\kappa_{m_l}$ (i.e., $\eta=2$).
The size of retrieved and generated training set $|\mathcal{D}^{\rm R}_l|=|\mathcal{D}^{\rm G}_l|=50$ for MAG-CS, and $|\mathcal{D}^{\rm R}_l|=|\mathcal{D}^{\rm G}_l|=100$ for Amazon.
For the CNN classifier \cite{kim2014convolutional}, following \cite{meng2018weakly}, we use filters with widths 2, 3, 4, and 5. For each width, we generate 20 feature maps. The maximum input sequence length is set to be 200 for both datasets. The CNN classifier is trained using SGD with the training batch size of 256.
The hyperparameter configuration of all baselines can be found in the Appendix.

\subsection{Performance Comparison}
\begin{table}[t]
\centering
\caption{Performance of compared methods on MAG-CS and Amazon. Bold: the highest score of weakly supervised methods. *: significantly worse than \textsc{\model} (p-value < 0.05). **: significantly worse than \textsc{\model} (p-value < 0.01).}
\vspace{-0.5em}
\scalebox{0.82}{
\begin{tabular}{@{}ccccc@{}}
\hline
\multicolumn{1}{c}{\multirow{2}{*}{\textbf{Algorithm}}} & \multicolumn{2}{c}{\textbf{MAG-CS}}                         & \multicolumn{2}{c}{\textbf{Amazon}}                         \\ \cline{2-5} 
\multicolumn{1}{c}{}                                 & \multicolumn{1}{c}{Micro-F1} & \multicolumn{1}{c}{Macro-F1} & \multicolumn{1}{c}{Micro-F1} & \multicolumn{1}{c}{Macro-F1} \\ \hline
WeSTClass \cite{meng2018weakly}       &  0.464**  &  0.326**  & 0.519**  &    0.547**    \\ 
WeSTClass+Metadata                    &  0.525**  &  0.369**  & 0.610**  &    0.603**    \\ 
MetaCat \cite{zhang2020minimally}     &  0.488**  &  0.403**  & 0.664**  &    0.657**    \\
META \cite{mekala2020meta}            &  0.398**  &  0.373**  & 0.664**  &    0.662      \\ 
LOTClass \cite{meng2020weakly}        & 0.124**   & 0.107**   & 0.658*   &    0.589**    \\ \hline
Metapath2Vec \cite{dong2017metapath2vec} & 0.436** & 0.414**  & 0.619**  &    0.611**    \\
HIN2Vec \cite{fu2017hin2vec}          & 0.408**   & 0.350**   & 0.628*   &    0.566**    \\ 
HGT \cite{hu2020heterogeneous}        & 0.151**   & 0.136**   & 0.272**  &    0.211**    \\ \hline
\textsc{\model}-NoHigherOrder         & 0.549*    & 0.476**   & 0.682    & \textbf{0.670}\\
\textsc{\model}-NoSpecificity         & 0.553*    & 0.499     & 0.675*   & 0.664         \\
\textsc{\model}                       & \textbf{0.571} & \textbf{0.501}  & \textbf{0.689} & \textbf{0.670} \\ \hline \hline
Supervised BERT \cite{devlin2019bert} & 0.798     & 0.717     & 0.952    & 0.952         \\ \hline
\end{tabular}
}
\vspace{-0.5em}
\label{tab:exp_results}
\end{table}

Table \ref{tab:exp_results} shows the Micro/Macro-F1 scores of compared algorithms. We repeat each experiment 5 times with the mean reported. To measure statistical significance, we conduct a two-tailed unpaired t-test to compare \textsc{\model} with each baseline approach. The significance level is also marked in Table \ref{tab:exp_results}.

From Table \ref{tab:exp_results}, we observe that: 
(1) \textsc{\model} consistently achieves the best performance. In most cases, the gap between \textsc{\model} and baselines is statistically significant. 
(2) The full \textsc{\model} model outperforms \textsc{\model}-NoHigherOrder and \textsc{\model}-NoSpecificity on both datasets (although not significant in some cases), which validates our claim that higher-order metadata information and metadata specificity are helpful to text classification. 
(3) Some weakly supervised text classification methods, such as WeSTClass+Metadata, MetaCat, and \textsc{\model}-NoHigherOrder, consider single metadata types only. The advantage of \textsc{\model} over these methods is larger on MAG-CS than it is on Amazon. This is possibly because higher-order motif structures can be better exploited in the MAG-CS network. Specifically, 4 out of 7 candidate motif patterns used in MAG-CS are higher-order, while only 1 out of 4 is higher-order in Amazon. 
(4) LOTClass is a strong baseline on Amazon but performs quite poorly on MAG-CS. This is because most category names in MAG-CS are phrases, and separating them into single words actually distorts the meaning of those categories.
(5) \textsc{\model} outperforms HIN embedding approaches (i.e., Metapath2Vec, HIN2Vec, and HGT) by a clear margin. We believe this is because the constructed HIN losses local context information of each term in the text. 
In contrast, \textsc{\model} models context proximity (i.e., $\mathcal{J}_{\rm Ctxt}$ in Eq. (\ref{eqn:local})) in addition to the HIN.

\subsection{Analysis of Higher-order Metadata}
\label{sec:high}
\begin{table}[t]
\centering
\caption{Proportion of each motif pattern in selected motif instances on MAG-CS. We show 10 (out of 20) categories. V: \textsc{Venue}, A: \textsc{Author}, Y: \textsc{Year}, T: \textsc{Term}.}
\vspace{-0.5em}
\scalebox{0.82}{
\begin{tabular}{c|ccccccc}
\hline
\textbf{Category}      & \textbf{V} & \textbf{A} & \textbf{T} & \textbf{V-Y} & \textbf{V-A} & \textbf{A-A} & \textbf{A-Y} \\ \hline
computer security      & 0          & 0.24       & 0.24       & 0.34         & 0.14         & 0.04         & 0            \\
computer vision        & 0          & 0.20       & 0.14       & 0.36         & 0.20         & 0.06         & 0.04         \\
data mining            & 0          & 0.24       & 0.18       & 0.36         & 0.22         & 0            & 0            \\
database               & 0          & 0.06       & 0.38       & 0.44         & 0.10         & 0.02         & 0            \\
embedded system        & 0          & 0.24       & 0.54       & 0            & 0.14         & 0.08         & 0            \\
information retrieval  & 0          & 0.22       & 0.02       & 0.44         & 0.32         & 0            & 0            \\
machine learning       & 0          & 0.30       & 0.38       & 0            & 0.16         & 0.06         & 0.10         \\
multimedia             & 0          & 0.10       & 0.06       & 0.48         & 0.34         & 0.02         & 0            \\
real time computing &	0.04 &	0.16 &	0.54 &	0.08 &	0.12 &	0.06 &	0 \\
theoretical comp. sci. & 0          & 0.42       & 0          & 0.54         & 0.04         & 0            & 0            \\ \hline
overall                  & 0.003      & 0.211      & 0.301      & 0.248        & 0.186        & 0.032        & 0.019        \\ \hline
\end{tabular}
}
\label{tab:proportion}
\end{table}

\begin{figure}[t]
\centering
\subfigure[MAG-CS]{
\includegraphics[width=0.23\textwidth]{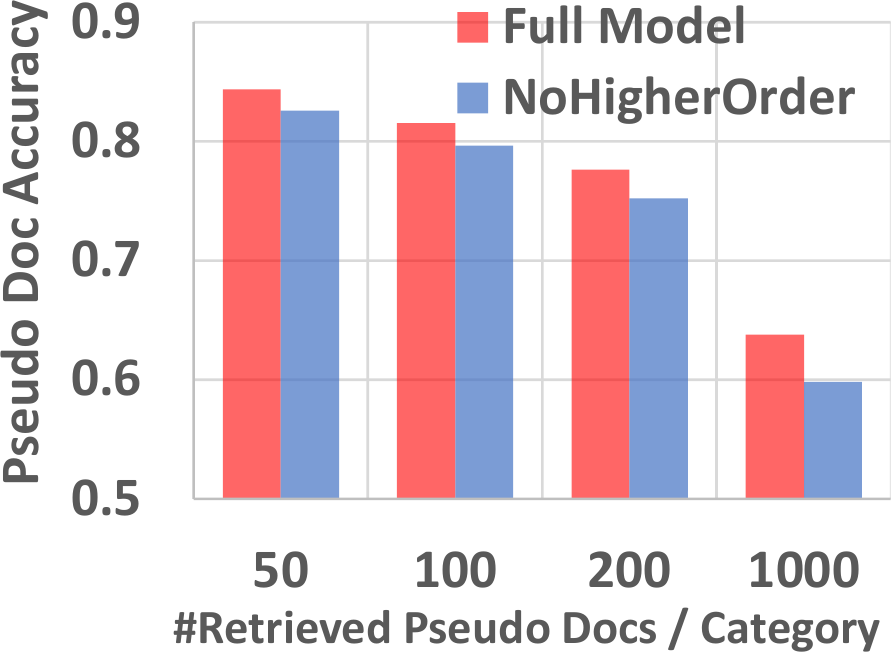}}
\subfigure[Amazon]{
\includegraphics[width=0.23\textwidth]{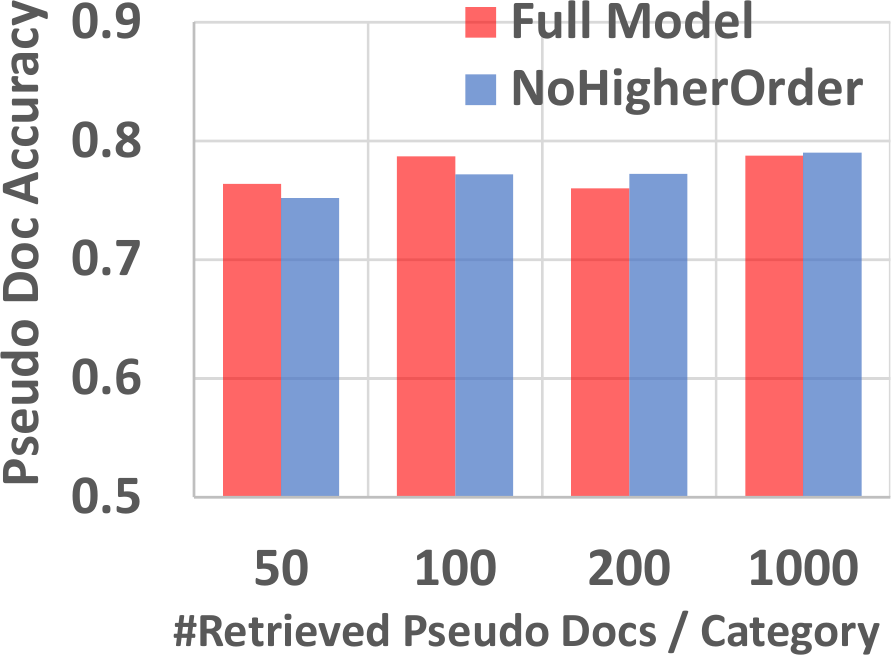}}
\vspace{-1.0em}
\caption{Accuracy of retrieved pseudo-labeled training documents with and without using higher-order metadata.} 
\vspace{-0.5em}
\label{fig:pseudo}
\end{figure}

The comparison between the full \textsc{\model} model and two ablation versions already show the positive contribution of both higher-order metadata and metadata specificity. In Sections \ref{sec:high} and \ref{sec:spec}, we would like to give more detailed analyses of these two factors. We start from higher-order metadata in this section.

\vspace{1mm}

\noindent \textbf{Observation 1: Higher-order instances cover a large proportion of selected instances.} Table \ref{tab:proportion} presents the proportion of each motif pattern in selected motif instances on MAG-CS. (For example, if we select $|\mathcal{M}_l|=50$ motif instances for a category $l$, and 10 of them are instances of \textsc{Venue}-\textsc{Year}, then the proportion of \textsc{Venue}-\textsc{Year} is $10/50=0.20$.) Due to space limit, we show 10 (out of 20) categories. We also show the overall proportion of each pattern across all 20 categories. 
It can be observed that: (1) The overall proportion of \textsc{Venue}-\textsc{Year}, \textsc{Venue}-\textsc{Author}, \textsc{Author}-\textsc{Author}, and \textsc{Author}-\textsc{Year} is 48.5\% in total. In other words, nearly half of the motif instances selected by \textsc{\model} are higher-order. 
(2) The same motif pattern can play very different roles in different categories. For example, for ``theoretical computer science'', the proportion of \textsc{Venue}-\textsc{Year} is more than 50\%. However, for ``embedded system'', \textsc{\model} does not pick any \textsc{Venue}-\textsc{Year} instance, possibly because conferences related to ``embedded system'' often have papers belonging to ``real-time computing'' as well. Besides, only ``computer vision'' and ``machine learning'' categories have \textsc{Author}-\textsc{Year} instances selected. This is possibly because most \textsc{Author}-\textsc{Year} instances are infrequent, but machine learning and computer vision researchers have relatively more papers per year on average, so that the corresponding \textsc{Author}-\textsc{Year} instances can pass the minimum frequency threshold.

\vspace{1mm}

\noindent \textbf{Observation 2: Higher-order instances improve the quality of retrieved pseudo training data.}
To explore the benefit of leveraging higher-order metadata signals in pseudo training data collection, we calculate the accuracies of pseudo-labeled training documents retrieved by the full \textsc{\model} model and \textsc{\model}-NoHigherOrder. (For example, if 1000 documents are retrieved in total, and for 800 of them, the pseudo label is the same as the true label, then the accuracy is $800/1000=0.80$.) Figure \ref{fig:pseudo} demonstrates the pseudo training data accuracy when we retrieve 50, 100, 200, and 1000 documents per category.

On MAG-CS, we can see the advantage of \textsc{\model} over \textsc{\model}-NoHigherOrder in retrieving pseudo training data. In fact, \textsc{\model} consistently outperforms \textsc{\model}-NoHigherOrder in terms of the retrieved training document accuracy. 
Intuitively, the accuracy of pseudo-labeled training data will affect the quality of the trained text classifier. By considering higher-order motif patterns, the full \textsc{\model} model is able to find more category-indicative instances and collect more accurate training samples, which can explain why it finally outperforms \textsc{\model}-NoHigherOrder in Table \ref{tab:exp_results}.
On Amazon, the gap between \textsc{\model} and \textsc{\model}-NoHigherOrder is not significant in terms of the retrieved training document accuracy, which is also reflected in their final classification performance in Table \ref{tab:exp_results}.

\subsection{Analysis of Specificity}
\label{sec:spec}
\begin{table*}
\centering
\caption{Motif instances close to each category at different specificity levels (from coarse to fine) on MAG-CS. Too general instances (in red) will not be selected by \textsc{\model}. $\kappa_{m_l}$: specificity of the category, $\kappa_m$: specificity of the motif instance.}
% \vspace{-0.5em}
\scalebox{0.82}{
\begin{tabular}{c|c|c|c}
\hline
               \textbf{Choice of} $\kappa_m$                                          &  \textbf{database} ($\kappa_{m_l} = 0.498$)                                                                                                                                                                          &  \textbf{data mining} ($\kappa_{m_l} = 0.588$)                                                                                                                                                                  &  \textbf{information retrieval} ($\kappa_{m_l} = 0.576$)                                                                                                                                                     \\ \hline
 \begin{tabular}[c]{@{}c@{}} $0 \leq \kappa_m < \kappa_{m_l}$ \\ \textbf{Not Selected} \end{tabular} & \leavevmode\color{red} \begin{tabular}[c]{@{}c@{}}\textsc{Term}[\textit{records}]\\  \textsc{Term}[\textit{index}]\\ \textsc{Venue}[\textit{CIKM}] \end{tabular}                                                                                   & \leavevmode\color{red} \begin{tabular}[c]{@{}c@{}}\textsc{Term}[\textit{mining}]\\ \textsc{Venue}[\textit{KDD}]\\ \textsc{Term}[\textit{big data}] \end{tabular}                                                                              & \leavevmode\color{red} \begin{tabular}[c]{@{}c@{}}\textsc{Term}[\textit{retrieval}]\\ \textsc{Term}[\textit{documents}]\\ \textsc{Venue}[\textit{CIKM}]\end{tabular}                                                                    \\ \hline
\begin{tabular}[c]{@{}c@{}} $\kappa_{m_l} \leq \kappa_m < 2\kappa_{m_l}$ \\ \textbf{Not Selected}\end{tabular} & \leavevmode\color{red} \begin{tabular}[c]{@{}c@{}}\textsc{Term}[\textit{sql}]\\ \textsc{Term}[\textit{relational database management system}]\\ \textsc{Venue}[\textit{SIGMOD}]\end{tabular}                                                  & \leavevmode\color{red} \begin{tabular}[c]{@{}c@{}}\textsc{Term}[\textit{knowledge extraction}]\\ \textsc{Term}[\textit{association rule learning}]\\ \textsc{Term}[\textit{data mining algorithm}] \end{tabular}                     & \leavevmode\color{red} \begin{tabular}[c]{@{}c@{}}\textsc{Venue}[\textit{SIGIR}]\\ \textsc{Term}[\textit{document retrieval}]\\ \textsc{Term}[\textit{text retrieval}] \end{tabular}                                          \\ \hline
$2\kappa_{m_l} \leq \kappa_m < 3\kappa_{m_l}$                                                      & \leavevmode\color{blue} \begin{tabular}[c]{@{}c@{}}\textsc{Term}[\textit{dbmss}]\\ \textsc{Term}[\textit{database research}]\\ \textsc{Venue}[\textit{SIGMOD}]-\textsc{Year}[\textit{2018}]\end{tabular}                                  & \leavevmode\color{blue} \begin{tabular}[c]{@{}c@{}}\textsc{Term}[\textit{apriori algorithm}]\\ \textsc{Venue}[\textit{KDD}]-\textsc{Year}[\textit{2008}]\\  \textsc{Author}[\textit{Jiawei Han}]\end{tabular}                                       & \leavevmode\color{blue} \begin{tabular}[c]{@{}c@{}}\textsc{Venue}[\textit{SIGIR}]-\textsc{Year}[\textit{2019}]\\ \textsc{Term}[\textit{ir evaluation}]\\ \textsc{Venue}[\textit{CLEF}]\end{tabular}                                                         \\ \hline
$3\kappa_{m_l} \leq \kappa_m < 4\kappa_{m_l}$                                                       & \leavevmode\color{blue} \begin{tabular}[c]{@{}c@{}}\textsc{Term}[\textit{sql database}]\\ \textsc{Venue}[\textit{VLDB}]-\textsc{Year}[\textit{2008}]\\ \textsc{Venue}[\textit{ICDE}]-\textsc{Author}[\textit{David B. Lomet}]\end{tabular}                 & \leavevmode\color{blue} \begin{tabular}[c]{@{}c@{}}\textsc{Venue}[\textit{KDD}]-\textsc{Year}[\textit{2007}]\\ \textsc{Venue}[\textit{KDD}]-\textsc{Author}[\textit{Usama M. Fayyad}]\\ \textsc{Venue}[\textit{KDD}]-\textsc{Author}[\textit{Mohammed Zaki}]\end{tabular} & \leavevmode\color{blue} \begin{tabular}[c]{@{}c@{}}\textsc{Venue}[\textit{SIGIR}]-\textsc{Year}[\textit{1994}]\\ \textsc{Author}[\textit{Donna Harman}]\\ \textsc{Term}[\textit{faceted search}]\end{tabular}                  \\ \hline
$4\kappa_{m_l} \leq \kappa_m$                                                      & \leavevmode\color{blue} \begin{tabular}[c]{@{}c@{}}\textsc{Venue}[\textit{VLDB}]-\textsc{Year}[\textit{1998}]\\ \textsc{Author}[\textit{H.-P. Kriegel}]-\textsc{Author}[\textit{Daniel A. Keim}]\\ \textsc{Venue}[\textit{VLDB}]-\textsc{Author}[\textit{Michael J. Carey}]\end{tabular} & \leavevmode\color{blue} \begin{tabular}[c]{@{}c@{}}\textsc{Venue}[\textit{KDD}]-\textsc{Year}[\textit{1996}]\\ \textsc{Venue}[\textit{KDD}]-\textsc{Author}[\textit{Heikki Mannila}]\\ \textsc{Venue}[\textit{KDD}]-\textsc{Author}[\textit{Charu C. Aggarwal}]\end{tabular}          & \leavevmode\color{blue} \begin{tabular}[c]{@{}c@{}}\textsc{Venue}[\textit{SIGIR}]-\textsc{Year}[\textit{2004}]\\ \textsc{Venue}[\textit{SIGIR}]-\textsc{Author}[\textit{Noriko Kando}]\\ \textsc{Venue}[\textit{SIGIR}]-\textsc{Author}[\textit{Nicholas J. Belkin}]\end{tabular} \\ \hline
\end{tabular}
}
\label{tab:specificity}
\end{table*}

Now we proceed to metadata specificity. To explain why considering specificity is important in motif instance selection, we list motif instances that are close to each category at different specificity levels in Table \ref{tab:specificity}. We choose three categories in MAG-CS -- ``database'', ``data mining'', and ``information retrieval''. Note that in the hyperparameter settings of \textsc{\model}, we require $\kappa_m \geq 2\kappa_{m_l}$. Therefore, instances in the first two rows (in red) will not be selected by \textsc{\model}. 

We have two findings from Table \ref{tab:specificity}. (1) When $\kappa_m$ is smaller, the instances have broader semantic coverage, meanwhile become less category-indicative. For example, \textsc{Venue}[\textit{CIKM}] is broader than the three categories because CIKM accepts papers from all these three areas. Although close to ``database'' and ``information retrieval'' in the embedding space, it should be filtered out since it is not discriminative enough to indicate either category. (2) When $\kappa_m$ becomes larger, more higher-order motif instances emerge. For example, when $2\kappa_{m_l} \leq \kappa_m < 3\kappa_{m_l}$, we start to see \textsc{Venue}-\textsc{Year} instances; when $3\kappa_{m_l} \leq \kappa_m < 4\kappa_{m_l}$, \textsc{Venue}-\textsc{Author} instances emerge; when $4\kappa_{m_l} \leq \kappa_m$, we can find \textsc{Author}-\textsc{Author} instances. Such metadata combinations express more accurate semantics, meanwhile cover fewer documents than single metadata instances. Overall, we believe setting $\eta=2$ can strike a good balance here. 

\begin{table}[t]
\caption{Running time (in hours) of weakly supervised text classification methods on the two datasets.}
\vspace{-0.5em}
\scalebox{0.82}{
\begin{tabular}{c|ccccc}
\hline
       & \textbf{WeSTClass} & \textbf{MetaCat} & \textbf{LOTClass} & \textbf{META} & \textbf{\textsc{\model}} \\ \hline
MAG-CS & 2.9       & 0.4     & 6.1      & 74.7 & 2.8        \\
Amazon & 0.2       & 0.3     & 1.1      & 11.0 & 2.1        \\ \hline
\end{tabular}
}
\label{tab:time}
\vspace{-0.5em}
\end{table}

\subsection{Efficiency}
Table \ref{tab:time} shows the running time of all weakly supervised text classification methods on Intel Xeon E5-2680 v2 @ 2.80GHz and one NVIDIA GeForce GTX 1080. Since \textsc{\model} considers higher-order network structures, its running time is longer than WeSTClass and MetaCat. However, compared with META which also leverages metadata combinations, \textsc{\model} is 26.6 times faster on MAG-CS and 5.2 times faster on Amazon.

\section{Related Work}
\noindent \textbf{Weakly Supervised Text Classification.}
Weakly supervised text classification aims to classify documents solely based on label surface names or category-indicative keywords. A pioneering approach is dataless classification \cite{chang2008importance,song2014dataless,yin2019benchmarking} which relies on Wikipedia to map labels and documents into the same semantic space and derive their relevance. Along another line, seed-guided topic models \cite{chen2015dataless,li2016effective} infer topics from descriptive keywords and predict labels from posterior category-topic assignments. 
Recently, neural models have been applied to weakly supervised text classification. \citet{meng2018weakly} propose to generate documents to train a neural classifier and refine the classifier via self-training. Their approach is further improved by introducing pre-trained language models. For example, ConWea \cite{mekala2020contextualized} utilizes contextualized word representations to detect category-indicative words for pseudo-label generation. LOTClass \cite{meng2020weakly} uses BERT to predict masked category names to find category-indicative keywords. X-Class \cite{wang2021x} leverages BERT representation of each word to cluster and align documents to categories. 

However, all these approaches consider only the text information and do not make use of metadata signals. Moreover, BERT-based approaches require the category names or keywords to be a single word in the vocabulary of BERT, while our \textsc{\model} framework can take phrases as category names.

\vspace{1mm}

\noindent \textbf{Metadata-Aware Text Classification.} There are many efforts to incorporate metadata into text classification in a specific domain. For example, \citet{tang2015learning} 
% and \citet{chen2016neural} 
consider user and product information in document-level sentiment analysis; \citet{zubiaga2017towards} and \citet{zhang2017rate} leverage user profile information for tweet geolocalization. To deal with the general metadata-aware text classification task, \citet{kim2019categorical} add categorical metadata representation into a neural classifier; \citet{zhang2021match} present a Transformer architecture to encode metadata. While achieving inspiring performance, these approaches are fully supervised and require massive annotated training data. In contrast, our method only requires label surface names as supervision.

Recently, \citet{zhang2020minimally,zhang2021hierarchical,zhang2021minimally,zhang2019higitclass} use a small set of labeled documents or keywords as supervision to categorize text with metadata. However, their methods consider each metadata instance separately and fail to model higher-order interactions between different types of metadata. \citet{mekala2020meta} adopt motif patterns to iteratively discover topic-related motif instances and retrieve pseudo-labeled training data. Compared with their method, \textsc{\model} is able to model the specificity of each motif instance, which is crucial when selecting category-indicative motif instances. 

\section{Conclusions and Future Work}
In this paper, we propose to study weakly supervised metadata-aware text classification from the HIN perspective, which avails us with additional higher-order network structures besides corpus. We identify the importance of modeling higher-order metadata information and metadata specificity. We then propose the \textsc{\model} framework that discovers indicative motif instances for each category to create pseudo-labeled training documents. Experimental results demonstrate the effectiveness of \textsc{\model} as well as the utility of considering higher-order metadata and specificity.

In the future, it is of interest to extend our framework to hierarchical or multi-label text classification, where each document can belong to more than one category. In this setting, categories are no longer mutually exclusive, and one needs to reconsider how to select representative motif instances and assign pseudo labels.
Besides, we would like to explore the potential of automatically detecting candidate motif patterns under weak supervision.

\section*{Acknowledgments}
We thank Frank F. Xu and Dheeraj Mekala for their help with the experimental setup and anonymous reviewers for their valuable and insightful feedback.
Research was supported in part by US DARPA KAIROS Program No.\ FA8750-19-2-1004, SocialSim Program No.\ W911NF-17-C-0099, and INCAS Program No.\ HR001121C0165, National Science Foundation IIS-19-56151, IIS-17-41317, and IIS 17-04532, and the Molecule Maker Lab Institute: An AI Research Institutes program supported by NSF under Award No.\ 2019897. Any opinions, findings, and conclusions or recommendations expressed herein are those of the authors and do not necessarily represent the views, either expressed or implied, of DARPA or the U.S. Government.

\bibliographystyle{ACM-Reference-Format}
\balance
\bibliography{wsdm}

\newpage
\nobalance

\appendix
\section{Appendix}
In the Appendix, we first show how we optimize the objective of our joint embedding and specificity learning step (i.e., Eq. (\ref{eqn:obj})). Then, we give more detailed information of the two datasets and all compared methods. Finally, we present additional experimental results including the effect of retrieval and generation strategies as well as the visualization of our joint embedding space.

\subsection{Optimization of the Objective}
\label{sec:app_opt}
To optimize Eq. (\ref{eqn:obj}), we adopt the negative sampling \cite{mikolov2013distributed} technique. Following \cite{tang2015pte}, each time, we alternately select one term from the objective. Taking $\mathcal{J}_{\rm Doc}$ as an example. We first randomly sample a motif instance $m$. Given $m$, we randomly sample a positive document $d$ (i.e., $m$ appears in $d$) and several negative documents $d'$ from $\mathcal{D}$ \footnote{Inspired by \cite{mikolov2013distributed}, the negative sample distribution $\propto {\rm \#motif}(d)^{3/4}$, where ${\rm \#motif}(d)$ is the number of motif instances appearing in $d$.}. Then, we need to optimize the following objective.
\begin{equation}
\mathcal{J}_{\rm Doc} = -\log \sigma(\kappa_m \bme_m^T \bme_d) - \sum_{d'} \sigma(-\kappa_m \bme_m^T \bme_{d'}) + \text{const.}
\end{equation}
Here, $\sigma(\cdot)$ is the sigmoid function. Given a parameter $\theta$ ($\theta$ can be $\bme_u$, $\bme_d$, $\bme_{d'}$ or $\kappa_m$), we have
\begin{equation}
\frac{\partial \mathcal{J}_{\rm Doc}}{\partial \theta} = \Big(\sigma(\kappa_m \bme_m^T \bme_d)-1\Big)\frac{\partial \kappa_m \bme_m^T \bme_d}{\partial \theta} - \sum_{d'} \sigma(\kappa_m \bme_m^T \bme_{d'}) \frac{\partial \kappa_m \bme_m^T \bme_{d'}}{\partial \theta}, \notag
\end{equation}
where
\begin{equation}
  \frac{\partial \kappa_m \bme_m^T \bme_d}{\partial \bme_m} = \kappa_m \bme_d,\ \ \ \frac{\partial \kappa_m \bme_m^T \bme_d}{\partial \bme_d} = \kappa_m \bme_m,\ \ \ \frac{\partial \kappa_m \bme_m^T \bme_d}{\partial \kappa_m} = \bme_m^T \bme_d.
\end{equation}
Knowing the gradient, we optimize each embedding vector and specificity using gradient descent. To satisfy the constraints, after each update, one can do $\bme \leftarrow \bme/||\bme||_2$ if the embedding is not on the unit sphere, and $\kappa \leftarrow 0$ if $\kappa < 0$.

\subsection{Datasets}
Table \ref{tab:categories} shows the surface names of 20 categories in MAG-CS and 10 categories in Amazon.

\begin{table}[t]
\caption{List of categories in MAG-CS and Amazon.}
\vspace{-0.5em}
\scalebox{0.82}{
\begin{tabular}{>{\centering\arraybackslash}p{1.7cm}|>{\centering\arraybackslash}p{4.75cm}|>{\centering\arraybackslash}p{2.45cm}} \hline
                  & \textbf{MAG-CS \cite{zhang2021match}}   & \textbf{Amazon \cite{mcauley2013hidden}}  \\ \hline
Categories & \begin{tabular}[c]{@{}c@{}}
information retrieval \\
computer hardware \\
programming language \\
theoretical computer science \\
speech recognition \\
real time computing \\
database \\
embedded system \\
multimedia \\
machine learning \\
natural language processing \\
software engineering \\
computer network \\
world wide web \\
computer security \\
computer graphics \\
parallel computing \\
data mining \\
human computer interaction \\
computer vision \\
\end{tabular} & 
\begin{tabular}[c]{@{}c@{}}
android \\
books \\
cd \\
clothing \\
electronics \\
health \\
kitchen \\
movies \\
sports \\
video games \\
\end{tabular} \\ \hline
\end{tabular}
}
\vspace{-0em}
\label{tab:categories}
\end{table}

\subsection{Compared Methods}
The code source and hyperparameter configuration of each compared method are explained below.

\subsubsection{WeSTClass and WeSTClass+Metadata \cite{meng2018weakly}} 
We run the code from the first author's GitHub\footnote{\url{https://github.com/yumeng5/WeSTClass}}. The maximum number of self-training iterations is changed from 5000 to 1000 as we find the performance starts to drop after 1000 iterations. When concatenating metadata and text together as the input to WeSTClass+Metadata, the order is \textsc{Author}, \textsc{Venue}, \textsc{Year}, text for MAG-CS and \textsc{User}, \textsc{Product}, text for Amazon. The maximum input sequence length is 200. Other hyperparameters are set by default.

\subsubsection{MetaCat \cite{zhang2020minimally}} 
We use the code from the first author's GitHub\footnote{\url{https://github.com/yuzhimanhua/MetaCat}}. MetaCat takes a small set of labeled documents, instead of category names, as supervision. To align the experiment setting, we use the pseudo-labeled documents retrieved by \textsc{\model} as the supervision of MetaCat. The maximum input sequence length is 200. The number of training epochs is 40. Other hyperparameters are by default.

\subsubsection{META \cite{mekala2020contextualized}} 
The code is from the first author's GitHub\footnote{\url{https://github.com/dheeraj7596/META}}. The input motif patterns of META are the same as those of our \textsc{\model} model. Instead of using their phrase mining code, we directly input our phrase chunking results into their model. All hyperparameters are by default.

\subsubsection{LOTClass \cite{meng2020weakly}}
The code is from the first author's GitHub\footnote{\url{https://github.com/yumeng5/LOTClass}}. LOTClass takes category names or keywords as supervision, but each category name/keyword must be a single word in the vocabulary of BERT. (Other BERT-based weakly supervised text classifiers \cite{mekala2020contextualized,wang2021x} also suffer from this problem.) To apply it to our datasets, we separate each phrase category name into single words and remove those common words that appear in multiple category names. The maximum input sequence length is 120. The keyword matching threshold is 10 for MAG-CS and 20 for Amazon. Other hyperparameters are by default. 

\begin{table}[t]
\caption{Meta-paths used by baselines on the two datasets.}
\vspace{-0.5em}
\scalebox{0.82}{
\begin{tabular}{>{\centering\arraybackslash}p{1.7cm}|>{\centering\arraybackslash}p{3.7cm}|>{\centering\arraybackslash}p{3.55cm}}
\hline
                  & \textbf{MAG-CS \cite{zhang2021match}}  & \textbf{Amazon \cite{mcauley2013hidden}} \\ \hline
Meta-paths & \begin{tabular}[c]{@{}c@{}}
\textsc{Venue$\rightarrow$Paper} \\
\textsc{Author$\rightarrow$Paper} \\
\textsc{Venue$\rightarrow$Paper$\rightarrow$Year} \\
\textsc{Author$\rightarrow$Paper$\rightarrow$Author} \\
\textsc{Venue$\rightarrow$Paper$\rightarrow$Author} \\
\textsc{Author$\rightarrow$Paper$\rightarrow$Year} \\
\textsc{Term$\rightarrow$Paper} \\
\end{tabular} & 
\begin{tabular}[c]{@{}c@{}}
\textsc{User$\rightarrow$Review} \\
\textsc{Product$\rightarrow$Review} \\
\textsc{User$\rightarrow$Review$\rightarrow$Product} \\
\textsc{Term$\rightarrow$Review} \\
\end{tabular} \\ \hline
\end{tabular}
}
\vspace{-0em}
\label{tab:paths}
\end{table}

\subsubsection{Metapath2Vec \cite{dong2017metapath2vec}, HIN2Vec \cite{fu2017hin2vec}, and HGT \cite{hu2020heterogeneous}}
Yang et al. \cite{yang2020heterogeneous} integrate 13 popular heterogeneous network representation learning models into one GitHub repository\footnote{\url{https://github.com/yangji9181/HNE}}, which contains the code of Metapath2Vec, HIN2Vec, and HGT. We use their code. Our constructed HIN has (\textsc{Document}, \textsc{Metadata}) and (\textsc{Document}, \textsc{Term}) edges. Metapath2Vec requires users to specify meta-paths. Thus, we view the motif patterns used by \textsc{\model} as meta-paths. Table \ref{tab:paths} shows the meta-paths used on the two datasets. For all three baselines, we change the embedding dimension from 50 to 100 for consistency with \textsc{\model}. For HGT, we change the number of training epochs from 100 to 500 as we observe higher performance. Other hyperparameters are by default.

\subsubsection{Supervised BERT \cite{devlin2019bert}}
We use the BertForSequenceClassification class\footnote{\url{https://huggingface.co/transformers/model_doc/bert.html\#bertforsequenceclassification}} from HuggingFace Transformers. The batch size is 16. The number of training epochs is 10. The model is optimized using AdamW with $lr=$ 5e--5. Other hyperparameters are by default.

\subsection{Additional Experiments: Effect of Retrieval and Generation}
We adopt two strategies to collect pseudo-labeled training data -- retrieval and generation. At the end of Section \ref{sec:method2}, we have already explained their respective merits. Now, we empirically show the advantage of combining these two strategies. In \textsc{\model}, we set $|\mathcal{D}^{\rm R}_l|=|\mathcal{D}^{\rm G}_l|=X$, where $X=50$ for MAG-CS and $X=100$ for Amazon. In other words, we collect $X$ retrieved pseudo training documents and $X$ generated ones for each category. We compare this strategy with four variants. Two of the variants do not generate any training data but collect $X$ and $2X$ retrieved documents, respectively, for each category. In contrast, the other two variants do not retrieve any training data but generate $X$ and $2X$ pseudo documents, respectively, for each category. Table \ref{tab:gen} compares the classification performance of \textsc{\model} and the four variants.

\begin{table}[t]
\centering
\caption{Effect of retrieval and generation strategies in creating pseudo-labeled training data. Bold: the highest score. *: significantly worse than \textsc{\model} (p-value < 0.05). **: significantly worse than \textsc{\model} (p-value < 0.01).}
\vspace{-0.5em}
\scalebox{0.82}{
\begin{tabular}{ccccc}
\hline
\multirow{2}{*}{\textbf{Algorithm}}       & \multicolumn{2}{c}{\textbf{MAG-CS}}         & \multicolumn{2}{c}{\textbf{Amazon}}         \\ \cline{2-5} 
                                       & Micro-F1    & Macro-F1    & Micro-F1    & Macro-F1    \\ \hline
\begin{tabular}[c]{@{}c@{}} \textsc{\model} \\ ($X$ retrieved $+$ $X$ generated) \end{tabular} & \textbf{0.571} & \textbf{0.501} & \textbf{0.689}                          &      \textbf{0.670} \\ \hline
$X$ retrieved $+$ $0$ generated              & 0.555          & 0.489*         & 0.657**         & 0.642*         \\
$2X$ retrieved $+$ $0$ generated             & 0.527**        & 0.469**        & 0.667**        & 0.662         \\
$0$ retrieved $+$ $X$ generated              & 0.491**        & 0.449**        & 0.614**        & 0.598**        \\
$0$ retrieved $+$ $2X$ generated             & 0.486**        & 0.452**        & 0.623**        & 0.610**        \\ \hline
\end{tabular}
}
\label{tab:gen}
\vspace{-0em}
\end{table}

As we can see from Table \ref{tab:gen}: (1) \textsc{\model} consistently outperforms the four variants on both datasets. In most cases, the gap is statistically significant. This validates our claim that retrieval and generation strategies are complementary to each other. (2) If we compare ``$X$ retrieved $+$ $0$ generated'' with ``$2X$ retrieved $+$ $0$ generated'', the former performs better on MAG-CS while the latter is better on Amazon. This observation can be explained by Figure \ref{fig:pseudo}. On MAG-CS, the accuracy of retrieved pseudo training documents drops significantly when $X$ becomes larger. On Amazon, the accuracy just slightly fluctuates as $X$ increases, thus the model can perform better when using more retrieved training data. (3) If we compare ``$0$ retrieved $+$ $X$ generated'' with ``$0$ retrieved $+$ $2X$ generated'', the latter is slightly better. This is because the quality of generated training data is not affected by $X$, as each document is sampled independently. Therefore, it is always better to have more generated training data. (4) In general, retrieval-only variants perform better than generation-only variants. 

\subsection{Additional Experiments: Embedding Visualization}
\begin{figure}[t]
\centering
\subfigure[MAG-CS]{
\includegraphics[width=0.45\textwidth]{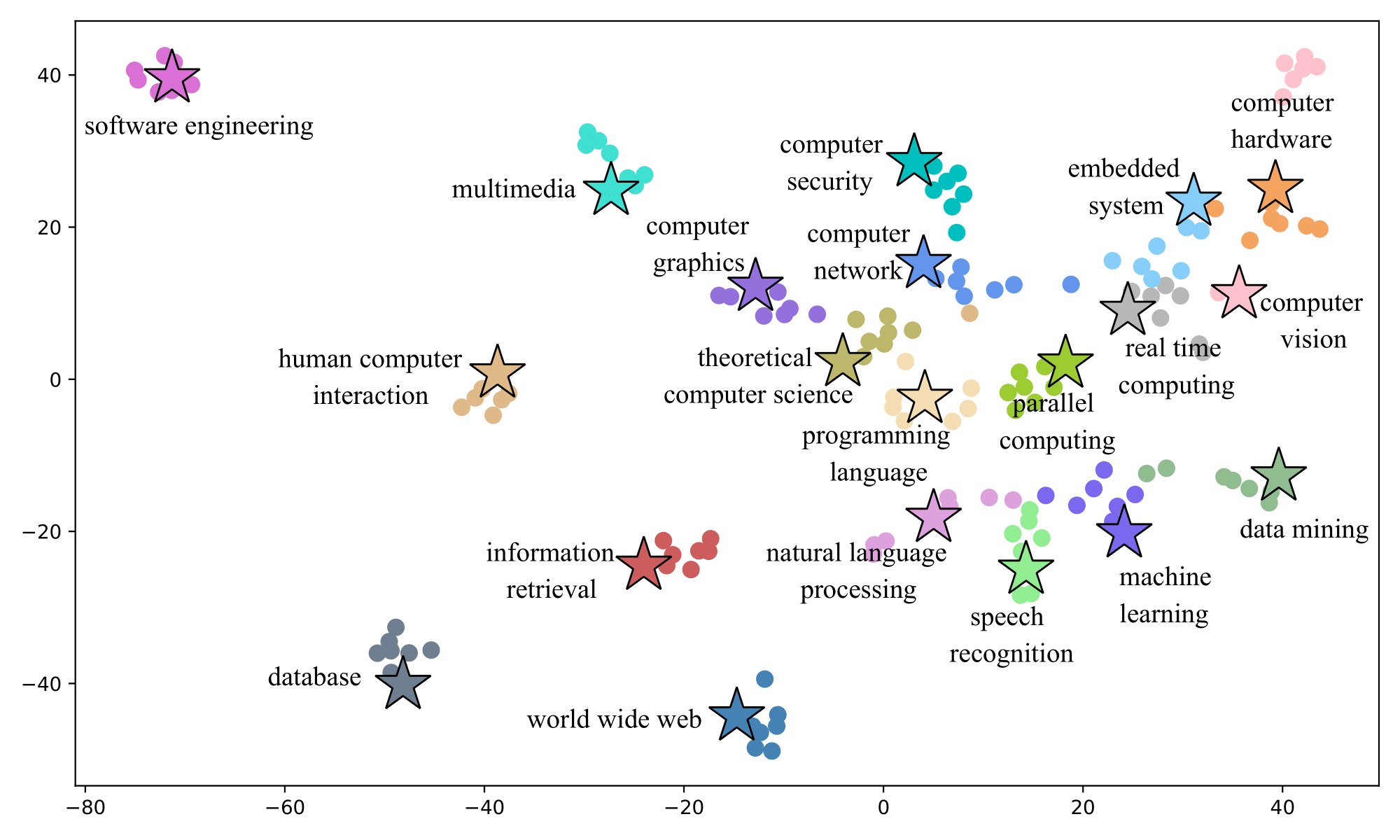}} \\
\subfigure[Amazon]{
\includegraphics[width=0.45\textwidth]{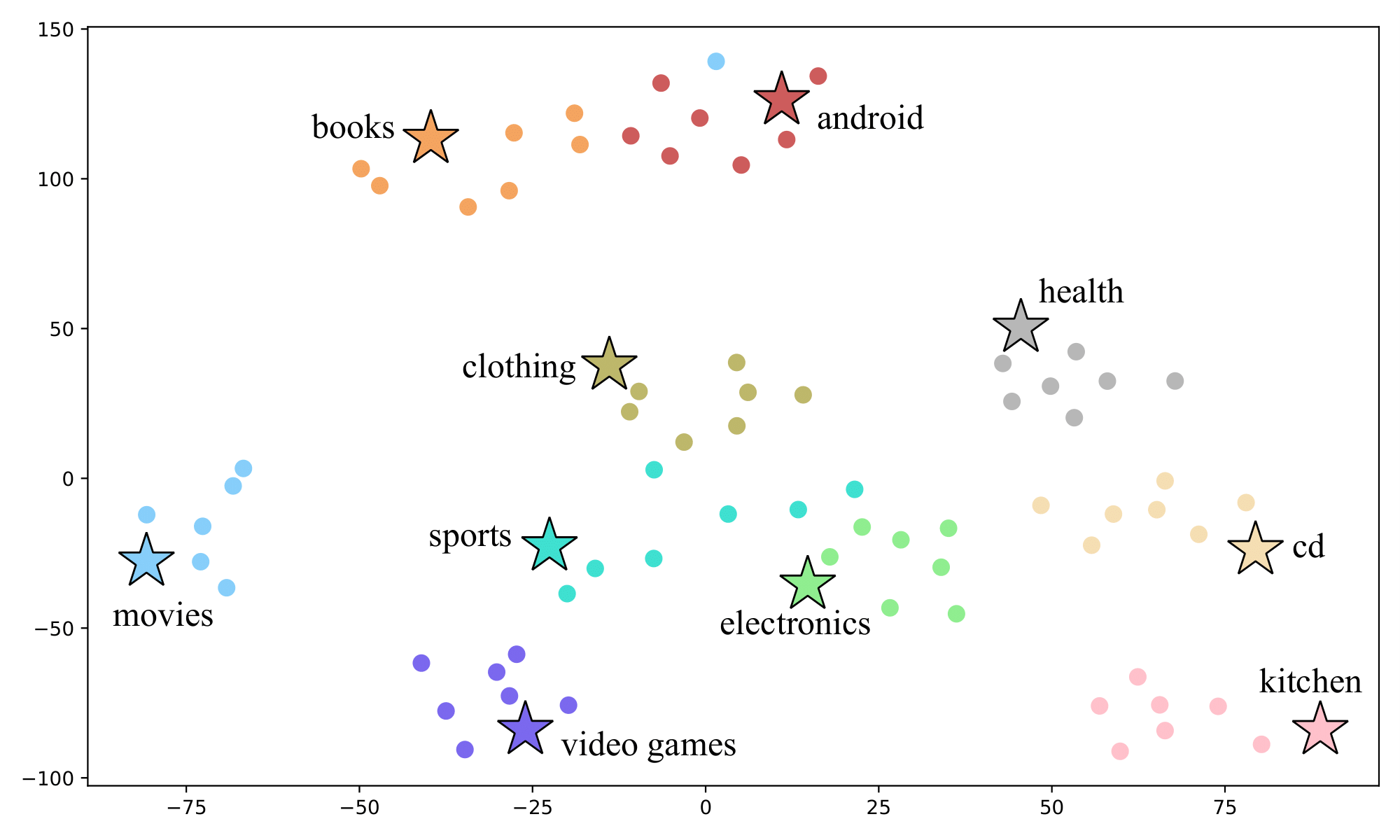}}
\vspace{-1.0em}
\caption{Embedding space visualization. Category name embeddings are denoted as stars, and the embeddings of selected motif instances are denoted as points with the same color as their corresponding categories.} 
\label{fig:visual}
\vspace{-0.5em}
\end{figure}

To reveal how categories and selected motif instances are distributed in our joint embedding space, we apply t-SNE \cite{maaten2008visualizing} to visualize their embeddings in Figure \ref{fig:visual}. Category name embeddings (i.e., $\{\bme_{m_l}: l \in \mathcal{L}\}$) are denoted as stars; embeddings of top-7 selected motif instances (i.e., $\{\bme_m: m \in \mathcal{M}_l\}$) are denoted as points with the same color as their corresponding categories. We observe that: (1) Selected motif instances surround their category names in most cases. (2) Semantically similar categories (e.g., ``data mining'' and ``machine learning'' in MAG-CS, ``clothing'' and ``sports'' in Amazon) are embedded closer.

\end{spacing}
\end{document}